\documentclass[]{interact}

\usepackage{epstopdf}
\usepackage{subfigure}

\usepackage[hidelinks]{hyperref}
\usepackage{amsmath,amssymb,amsfonts}
\usepackage{float}
\usepackage{algorithmic}
\usepackage{graphicx}
\usepackage{multirow}
\usepackage{textcomp}
\usepackage{bbm}
\usepackage[numbers]{natbib}
\usepackage{xcolor}
\usepackage{pgfplots}
\usepgfplotslibrary{fillbetween}
\usepackage{tikz}
\usetikzlibrary{arrows, decorations.markings}
\usetikzlibrary{chains, positioning}
\usepackage{dblfloatfix}

\DeclareMathOperator*{\argmin}{argmin}

\usepackage{natbib}
\bibpunct[, ]{(}{)}{;}{a}{}{,}
\renewcommand\bibfont{\fontsize{10}{12}\selectfont}

\theoremstyle{plain}

\theoremstyle{definition}

\theoremstyle{remark}

\begin{document}


\title{Data-Driven Probabilistic Energy Consumption Estimation for Battery Electric Vehicles with Model Uncertainty}

\author{
\name{Ayan Maity\textsuperscript{a} \thanks{Corresponding Author : Ayan Maity. Email : ayanmaity201@kgpian.iitkgp.ac.in} \thanks{CONTACT Ayan Maity. Email : ayanmaity201@kgpian.iitkgp.ac.in} and Sudeshna Sarkar\textsuperscript{b} \thanks{CONTACT Sudeshna Sarkar. Email : sudeshna@cse.iitkgp.ac.in}}
\affil{\textsuperscript{a}Department of Computer Science and Engineering, Indian Institute of Technology Kharagpur, Kharagpur, India; \textsuperscript{b}Department of Computer Science and Engineering, Indian Institute of Technology Kharagpur, Kharagpur, India}
}

\maketitle

\begin{abstract}
This paper presents a novel probabilistic data-driven approach to trip-level energy consumption estimation of battery electric vehicles (BEVs). As there are very few electric vehicle (EV) charging stations, EV trip energy consumption estimation can make EV routing and charging planning easier for drivers. In this research article, we propose a new driver behaviour-centric EV energy consumption estimation model using probabilistic neural networks with model uncertainty. By incorporating model uncertainty into neural networks, we have created an ensemble of neural networks using Monte Carlo approximation. Our method comprehensively considers various vehicle dynamics, driver behaviour and environmental factors to estimate EV energy consumption for a given trip. We propose relative positive acceleration (RPA), average acceleration and average deceleration as driver behaviour factors in EV energy consumption estimation and this paper shows that the use of these driver behaviour features improves the accuracy of the EV energy consumption model significantly. Instead of predicting a single-point estimate for EV trip energy consumption, this proposed method predicts a probability distribution for the EV trip energy consumption. The experimental results of our approach show that our proposed probabilistic neural network with weight uncertainty achieves a mean absolute percentage error of 9.3\% and outperforms other existing EV energy consumption models in terms of accuracy.
\end{abstract}


\section{Introduction}

Battery electric vehicles (BEVs) have emerged as promising replacements for internal combustion engine vehicles (ICEVs) in recent times. Electric Vehicles are more energy-efficient and eco-friendly than ICEVs. With the advent of machine learning (ML), neural network (NN), and big data, researchers and automotive engineers are taking interest in data-driven models for electric vehicle (EV) trip energy consumption prediction. 

This paper proposes a probabilistic data-driven approach for EV trip-level energy consumption estimation in this paper. In EV trip-level energy consumption estimation problem, the main objective is to predict the EV energy consumption for a given trip using various vehicle dynamics, environmental and driver behaviour factors along the trip route. Though EVs are gaining popularity, one of the main obstacles to mass adoption of EVs is range anxiety. As there are very few electric charging stations and EV charging time is much higher compared to conventional vehicles \citep{MONTOYA201787}, the EV might run out of energy in the middle of a trip and this gives rise to driver range anxiety \citep{evrp_range_anxiety}. Therefore, a reliable and accurate EV trip energy consumption estimation can help EV users plan their trip routes, and charging stops accordingly. It is important for EV users to get an estimate of the energy consumption for a given trip before the start of the journey. As EVs have a limited range, trip energy consumption estimation enables the EV users to compute the remaining range of the EV and schedule the recharging. Thus, the EV trip energy consumption model can reduce driver range anxiety \citep{cauwer_2018}. Additionally, EV energy consumption estimation helps in the optimal placement and deployment of charging stations. Accurate energy consumption prediction of EV trips can be useful in computing the energy consumption of an EV fleet in a local area, which is vital in building an optimal charging stations network \citep{fazeli2020two,yang2020data,ahmad2022optimal}. The EV energy estimation system can also be utilised in developing an optimal charging scheduling system for EV fleets \citep{korkas2017adaptive}.

 Researchers in the past have proposed EV energy consumption systems based on physical and mathematical modelling of EVs \citep{sarrafan_accurate_2017,prins2013electric}. But it is very difficult to quantify and incorporate driver behaviour and environmental factors into mathematical energy consumption models. Driver behaviour \citep{vatanparvar_extended_2019,driving_2021} and external environmental factors \citep{younes_analysis_2013} can have huge influence on energy use in a given trip. Data-driven methods for EV trip energy consumption have also been proposed in recent years \citep{cauwer_2018,qi_data-driven_2018,petkevicius_probabilistic_2021}. Though the data-driven methods generally perform better than the mathematical models, most of the previously proposed data-driven methods have not utilised driver behaviour factors effectively to estimate EV trip energy consumption. 

Electric vehicle energy consumption is susceptible to various internal, environmental, physical and driver-related factors \citep{younes_analysis_2013}. Even a minimal change or deviation in any of those factors can drastically alter the energy consumption of EVs. As the energy consumption is highly uncertain, deterministic models are not very reliable under real-world circumstances. It can be helpful for the users if the estimation method can predict a probability distribution or a confidence interval for the trip energy consumption. The energy consumption uncertainty will help the users to decide if charging stops are required in a trip. The energy consumption uncertainty and confidence interval can be utilised to develop robust and adaptive EV energy management systems and EV routing planning systems \citep{basso_electric_2021,basso2022dynamic}. The confidence interval will also be useful in building optimal charging stations networks. In order to address the problem of variability in energy consumption, probabilistic models \citep{petkevicius_probabilistic_2021,basso_electric_2021,jiang2023trip} 
have been proposed in various research studies. However, these probabilistic methods have not included model uncertainty or weight uncertainty in the models.

In this research study, we have proposed a probabilistic neural network (NN) with weight uncertainty-based method for EV trip energy consumption prediction. we have explored how driver behaviour along with other factors can be effectively integrated into EV energy consumption model.

The main contributions of this paper are as follows :
\begin{itemize}
    \item A novel driver behaviour-centric EV trip energy consumption model is developed, which considers relative positive acceleration (RPA), average acceleration and average deceleration to characterize driver behaviour.
    \item We have proposed a probabilistic neural network with weight uncertainty for EV trip energy consumption to capture the uncertainties involved in the energy estimation more accurately. By incorporating weight uncertainty, we have created an ensemble of neural networks by Monte Carlo approximation to predict probability distribution for EV trip energy consumption.
    \item We have thoroughly analysed the impacts of different vehicle dynamics, environmental and driver-related factors on EV energy consumption using our proposed method.
\end{itemize}

Section 2 presents the EV energy consumption problem statement. Section 3 discusses the related work on EV energy estimation, followed by different factors affecting EV energy consumption in Section 4. Section 5 describes the probabilistic neural network with model uncertainty model for energy estimation. Section 6 discusses the experimental results. Section 7 discusses the limitations of our approach and section 8 presents the conclusion and future work.

\section{EV Energy Consumption Problem Statement}
Given a trip and various vehicle dynamics, environmental and driver-related  factors along the trip route, the problem is to estimate the Electric Vehicle energy consumption for that trip.
    Energy Consumption for a trip is modelled as a function of trip route, speed profile, acceleration profile and other physical and environmental factors :
    \begin{equation}
    E = f(r, \hat{v}, \hat{a}, d, p)
    \end{equation} 
    where 
    $E$ is \textit{energy consumption for the trip},
    $r$ is \textit{trip route},
    $\hat{v}$ is \textit{speed profile},
    $\hat{a}$ is \textit{acceleration profile},
    $d$ is \textit{driver behaviour factors} and
    $p$ is \textit{other physical and environmental factors}. 

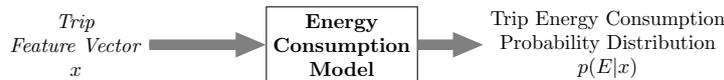
\begin{figure}[h!]
\centering
\scalebox{0.7}{
\begin{tikzpicture}[
roundnode/.style={circle, draw=green!60, fill=green!5, very thick, minimum size=7mm},
squarednode/.style={rectangle, draw=red!60, fill=red!5, very thick, minimum size=2cm},
newnode/.style={rectangle}
]
\tikzstyle{vecArrow} = [thick, decoration={markings,mark=at position
   1 with {\arrow[semithick]{triangle 60}}},
   double distance=1.4pt, shorten >= 5.5pt,
   preaction = {decorate},
   postaction = {draw,line width=1.4pt, black,shorten >= 4.5pt}]
\tikzstyle{myarrows}=[line width=1mm,draw=gray,-triangle 45,postaction={draw,line width=2mm, shorten >=4mm, -}]
\node [align=center] at (0,0) (1) {\textit{Trip} \\ \textit{Feature Vector} \\ $x$};

\node [draw=black, align=center] at (5,0) (2) {\textbf{Energy} \\ \textbf{Consumption} \\ \textbf{Model}};
\node [align=center] at (10,0) (3) {
Trip Energy Consumption \\ Probability Distribution\\$p(E|x)$};
\draw[myarrows] (1) to (2);
\draw[myarrows] (2) to (3);  
\end{tikzpicture}}
\caption{Probabilistic Energy Consumption Model}
\label{energy_model}
\end{figure}

As EV energy consumption depends on several uncertain and highly variable factors such as environmental factors, driver behaviour and vehicle speed profile \citep{younes_analysis_2013,petkevicius_probabilistic_2021,jiang2023trip}, we propose a probabilistic data-driven EV energy consumption estimation model, which can predict the energy consumption uncertainty along with the expected energy consumption. Our Objective is to predict EV energy consumption probability distribution for a trip based on various trip features such as speed profile, distance, environmental factors and driver-related factors, as shown in Figure \ref{energy_model}.

\section{Related Work}
In recent years, there has been a substantial increase in EV trip energy consumption research. EV energy consumption methods can be grouped into two major categories: mathematical models and data-driven models. Both mathematical and data-driven energy consumption methods have been extensively studied by researchers in the recent past. 

\subsection{Mathematical or Analytical Models}

Mathematical or analytical models generally compute EV trip energy consumption using Newton's laws of motion. \cite{prins2013electric} proposed a mathematical model that computes the EV trip energy consumption using various forces acting on the vehicle. They considered rolling resistance, frictional force, gravitational force and force due to acceleration to calculate the trip energy consumption. In \cite{tannahill2016driver}, the authors developed a driver alerting system, which is based on a mathematical model for EV energy consumption estimation. Their proposed EV energy consumption model incorporated various forces acting on the vehicle, wind speed, auxiliary energy expenditures and traffic condition. \cite{sarrafan_accurate_2017} created mathematical frameworks for predicting energy consumption for a given trip. This model calculates EV traction power using kinetic energy change, potential energy change, aerodynamic work, rolling resistance work. They have also included different internal power losses, energy regeneration efficiency, motor efficiency and traffic condition to accurately estimate the trip energy consumption. \cite{miri2021electric} came up with a mathematical EV energy consumption model using a simplified proportional-integral-differential controller-based driver model. \cite{alateef_api} used map services and routing API to collect vehicle speed profile data and generate a driving cycle for a given route. Then they used the generated driving cycle to predict EV energy consumption based on an analytical model. However, the major drawback of mathematical models is that these models can not effectively include driver behaviour and environmental factors making the mathematical models less accurate.

\renewcommand{\arraystretch}{1.5}

\begin{table}[h!]
\tbl{Literature Review of Different Data-Driven EV Energy Consumption Methods}
{\scalebox{0.8}{\begin{tabular}{p{40pt}p{85pt}p{100pt}p{115pt}p{80pt}}
\hline
\textbf{Article}                                                    & \textbf{Vehicle Dynamics Factors}                                                           & \textbf{Driver Behaviour Factors}                                                                     & \textbf{Environmental Factors}                                                                                        & \textbf{Methods}                                                                  \\ \hline
\cite{de_cauwer_energy_2015}                                                 & Speed, Constant Motion Factor              & -                                                                                                     & Positive and Negative Elevation Change, Temperature                      & Regression                                                                        \\ \hline
\cite{zheng_hybrid_2016}                                               & Speed, Acceleration                                                                         & -                                                                                                     & Elevation, Temperature                                                                                                & Regression Tree - Self Organizing Maps \\ \hline
\cite{qi_data-driven_2018}                                                    & Positive Kinetic Energy, Negative Kinetic Energy & -                                                                                                     & -                                                                                                                     & Regression and Multi-Layer Perceptron   \\ \hline
\cite{fukushima2018prediction} & Average Speed & - & Accumulation of up incline, Accumulation down incline, Temperature & Multiple Regression and Transfer Learning \\ \hline
\cite{cauwer_2018}                                                & Constant Motion Factor                                                                      & -                                                                                                     & Positive and Negative Elevation Change, Temperature, Aerodynamic Factor & Neural Network                                                                    \\ \hline
\cite{zhang2020energy}                                               & Speed                                                                                       & 95\% quantile of Acceleration, 5\% quantile of Deceleration & Temperature, Day of the week,  Time of the day                              & XGBoost                                                                           \\ \hline
\cite{basso_electric_2021}                                                  & Mass, Speed                                                                                 & -                                                                                                     & -                                                                                                                     & Bayesian Machine Learning             \\ \hline
\cite{petkevicius_probabilistic_2021} & Speed                                                                                       & -                                                                                                     & Altitude, Temperature, Wind Speed, Weekend, Road Type, Road Condition   & Probabilistic Neural Network       \\ \hline
\cite{ullah2021electric}                                                 & Speed                                                                                       & -                                                                                                     & Elevation, Time of the day                                                                                           & Ensemble Stacked Generalization \\ \hline
\cite{li2021prediction} & Speed & Acceleration, Deceleration & Time of day, Day of week, Temperature, Wet-dry condition & Stochastic Random Forest \\ \hline 
\cite{jiang2023trip} & Speed, Physical-based Energy Consumption & - & Time of day, Day of week, Traffic condition, Temperature, Accumulated mileage & Markov-based Gaussian Process Regression \\ \hline
    \end{tabular}}}
\label{related}
\end{table}

\subsection{Data-Driven Models}

Data-Driven EV energy consumption models are becoming popular in recent times due to the advancements in Big Data, Machine Learning (ML) and Deep Learning (DL) technologies. Table \ref{related} lists some recent research studies on data-driven EV energy consumption prediction methods. Table \ref{related} also shows different features that those methods have used. \cite{de_cauwer_energy_2015} proposed regression models by creating complex features to capture the effect of weather, driver behaviour and elevation change. They used speed and constant motion factor (CMF) as vehicle dynamic factors and they also included positive elevation, negative elevation and temperature as environmental input features to the Linear Regression model for predicting energy use. \cite{cauwer_2018} proposed a neural network model for predicting constant motion factor and aerodynamic factor and  they used those features along with other environmental factors to predict energy. 

\cite{diaz_alvarez_modeling_2014}  used velocity, positive acceleration, negative acceleration, and jerk to capture driver behaviour and used multilayer perceptron (MLP) regression to estimate EV energy consumption. However, they did not consider environmental factors. \cite{zheng_hybrid_2016} proposed a hybrid ML algorithm using self organizing maps (SOM) and regression tree. Although the model showed significant improvement in accuracy, this hybrid model lacks the effect of driver behaviour. \cite{liu2016modelling} used a multilevel mixed-effects linear regression model to esimate the EV trip energy consumption. They have also investigated the impacts of driving heterogeneity in EV energy consumption. \cite{bi2018residual} proposed a radial basis function neural network for EV remaining range estimation. They used battery state of charge (SoC), battery voltage, battery current and vehicle speed as input features. In \cite{sun_machine_2019}, the authors presented a gradient boosting decision tree method for EV range estimation using battery SoC, battery voltage, battery temperature, vehicle speed, visibility and precipitation as input features.

\cite{qi_data-driven_2018} presented a regression and multilayer perceptron (MLP) data-driven model which takes only positive kinetic energy and negative kinetic energy as input features to predict EV trip energy consumption. Their proposed model also doesn't consider the impact of any environmental and driver-related factors. \cite{fukushima2018prediction} developed  a multiple regression method for EV energy consumption and used transfer learning method to construct prediction models for new EV models. \cite{vatanparvar_extended_2019} proposed a data-driven model to predict future speed of the EV using nonlinear autoregressive model with eXogenous inputs (NARX) method. In \cite{zhao2022structured}, the authors developed a long short-term memory (LSTM) neural network with attention to accurately predict vehicle speed based on historical traffic flow data. \cite{ullah2021electric} proposed an ensemble stacked generalization (ESG) method for EV energy consumption rate prediction and they have shown that ESG method is more accurate than other conventional ML methods. The ESG method takes trip distance, average trip speed, auxiliary energy losses and road elevation to predict the trip energy consumption. This method did not include temperature and driver behaviour features for energy estimation. \cite{li2021prediction} constructed a two-step approach consisting of stochastic random forest (RF) and k nearest neighbour RF to predict energy consumption for Electric Buses. Their results showed that stochastic RF method had a MAPE error of 11.8\%. However, none of the above mentioned methods estimate EV trip energy consumption uncertainty, which can be very helpful in scheduling recharging stops and reducing user range anxiety.

Probabilistic models are yet to gain much popularity in EV energy estimation. \cite{petkevicius_probabilistic_2021} proposed a probabilistic deep learning model for EV energy use prediction. Though they created and trained probabilistic deep neural networks, they did not consider weight uncertainty of neural networks, which has the potential to improve the accuracy of the model. \cite{petkevicius_probabilistic_2021} used speed and different environmental factors to estimate energy use. Although they did not consider driver behaviour features and weight uncertainty in NNs. \cite{basso_electric_2021} came up with a bayesian machine learning model for energy prediction with uncertainty quantification. Their probabilistic model only uses mass and speed of the EV to predict energy use. \cite{basso2022dynamic} utilised the probabilistic EV energy consumption model developed in \cite{basso_electric_2021} to create a EV routing and recharging planner using safe reinforcement learning. \cite{jiang2023trip} developed a Markov-based speed profile generation and Gaussian process regression (GPR) method to predict trip-level energy consumption. However, \cite{petkevicius_probabilistic_2021}, \cite{basso_electric_2021}, and \cite{jiang2023trip} did not consider driver behaviour factors and model uncertainty in their EV energy consumption models.

With the introduction of vehicle-to-grid (V2G) technology, \cite{wan2018model} and \cite{li2019constrained} have proposed reinforcement learning-based EV charging-discharging scheduling methods. An adaptive dynamic programming-based optimal charging scheduling method has been proposed by \cite{korkas2017adaptive}. These methods did not use data-driven methods for estimating EV remaining state-of-charge (SoC). Data-driven EV charging stations load forecasting methods have also been developed \citep{zhou2022using,liu2022data}. \cite{yang2020data} proposed a data-driven method for optimal placement of EV charging stations.

Existing EV trip energy consumption methods do not comprehensively include various dynamics, environmental and driver behaviour factors. In this research work, we analyze different vehicle dynamics, environmental and driver behaviour features that can affect energy consumption for EVs. We incorporated model uncertainty or weight uncertainty in the probabilistic neural network, which enables us to create an ensemble of neural networks with the use of Monte Carlo approximation to estimate trip energy usage.

\section{Factors Affecting EV Energy Consumption}
\label{ev_model}
In this section, we introduce the EV trip data and then we have discussed different factors that affect EV energy consumption.

\subsection{Data}
We have used ChargeCar EV trip dataset \citep{chargecar_data}. CREATE Lab collected vehicle trip data as part of the ChargeCar project in Carnegie Mellon University. They collected energy consumption data for 423 trips, which included 50 EV trips. For every second of each trip they collected the following information :
\begin{itemize}
    \setlength\itemsep{0em}
    \item Speed ($m/s$)
    \item Acceleration ($m/s^2$) 
    \item Elevation Change ($m$)
    \item Distance Travelled ($m$)
    \item Power Consumption ($watt$)
    \item Ambient Temperature ($^\circ F$)
\end{itemize}

We have divided the actual EV trips into 5000 micro-trips using 3 levels of randomness \citep{zheng_hybrid_2016} in order to generate more training data. A trip is randomly picked from the dataset. Then we take a random starting point in the trip and cut out a random length micro-trip from the complete trip. In this way, we are able to create more data points, which is very essential for training complex data-driven models.

\subsection{Different Factors in Energy Consumption}
\label{diff_fac_sub}
There are different types of factors that impact EV energy consumption in different ways \citep{younes_analysis_2013}. We can group the contributing factors in 3 different categories : Vehicle Dynamics Factors, Driver Behaviour Factors, and Environmental Factors.

\subsubsection{Vehicle Dynamics Factors}
\label{vd_factor}

\begin{figure}[h!]
\begin{center}
    \subfigure[Energy Consumption Rate for different Average Speed values]{\includegraphics[width=7cm]{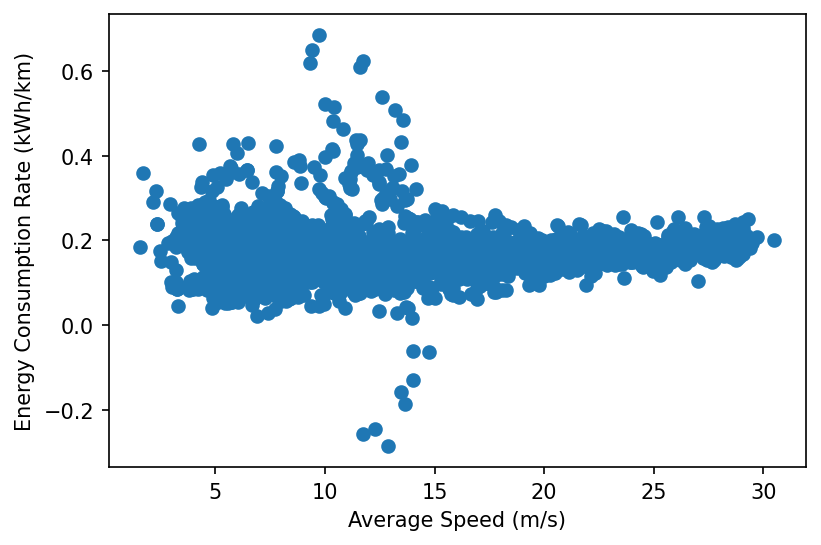}\label{avg_speed_fig}}
    \subfigure[Energy Consumption for different distances]{\includegraphics[width=7cm]{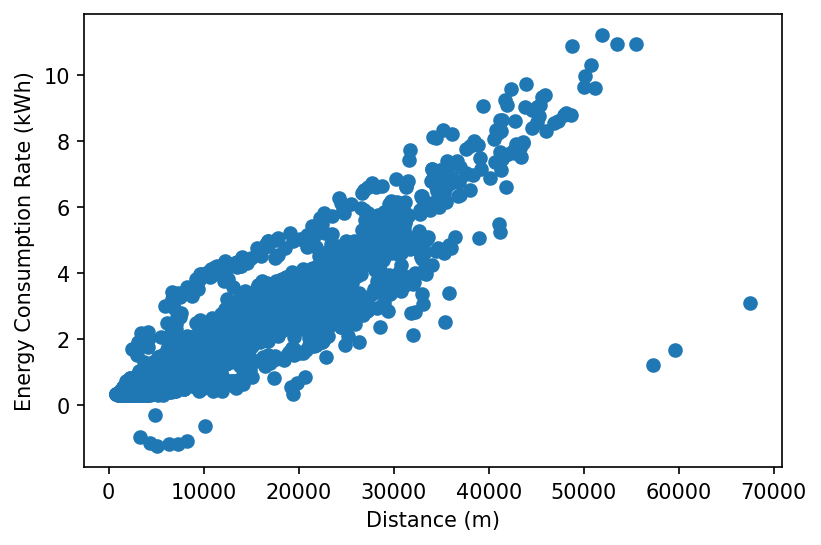}\label{dist_fig}}
    \caption{Energy Consumption for different distance and Average Speed values}
    \label{speed_dist_fig}
\end{center}
\end{figure}
Dynamics of the vehicle is essential in estimating trip energy usage \citep{zhang2020energy}.
Vehicle Dynamics factors can be captured by the speed profile for the trip. Speed Profile mostly depends on traffic along the trip route. In our model, we used Average Speed and Standard Deviation of Speed as Vehicle Dynamics Factors. In Figure \ref{avg_speed_fig}, Energy Consumption Rates (ECR) for different average speed values are plotted. The uncertainty and volatility of the energy consumption rate for smaller trips can be observed from Figure \ref{avg_speed_fig}. Figure \ref{dist_fig} illustrates the relation between energy consumption and trip distance. Energy consumption generally increases with distance. 
\subsubsection{Driver Behaviour Factors}
\label{db_factor}

\begin{figure}[h!]
    \subfigure[Energy Consumption Rate for different RPA values]{\scalebox{0.8}{
    \begin{tikzpicture}
    \begin{axis}[xticklabel style={
        /pgf/number format/fixed,
        /pgf/number format/precision=2
},
scaled y ticks=false, ylabel near ticks,xlabel=Relative Postive Acceleration ($m/s^2$),ylabel=Energy Consumption Rate (kwh/km)]
        \addplot[
                scatter, only marks]
         table[x=c1,y expr=\thisrow{c2}/3.6, col sep=comma]
            {Data/rpa_analysis.txt};
    \end{axis}
    \end{tikzpicture}\label{rpa_fig}}} 
    \subfigure[Energy Consumption Rate for different Average Acceleration values]{\scalebox{0.8}{\begin{tikzpicture}
    \begin{axis}[xticklabel style={
        /pgf/number format/fixed,
        /pgf/number format/precision=2
},
scaled y ticks=false,ylabel near ticks,xlabel=Average Acceleration ($m/s^2$) ,ylabel=Energy Consumption Rate (kwh/km)]
        \addplot[
                scatter, only marks]
         table[x=c1,y expr=\thisrow{c2}/3.6, col sep=comma]
            {Data/acc_analysis.txt};
    \end{axis}
    \end{tikzpicture}}\label{acc_fig}}
    \caption{Energy Consumption Rate for different RPA and Average Acceleration values}
    \label{db_analysis}
\end{figure}
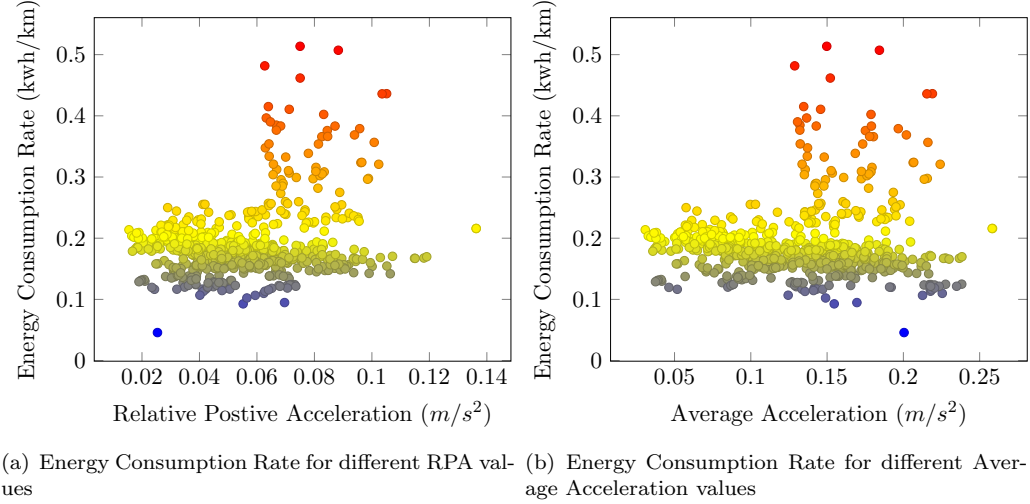

Driver behaviour (DB) can have a huge impact in the energy consumption of the EV. A relatively aggressive driver accelerates and decelerates very frequently and the EV consumes more energy as a result of this aggressive driving. Negative acceleration or deceleration can result in energy regeneration due to regenerative braking phenomenon in EVs \citep{qi_data-driven_2018}. So, it will be beneficial to have positive acceleration and negative acceleration (deceleration) as two different features in energy prediction. Figure \ref{acc_fig} shows energy consumption rate values and average acceleration values for different trips. It can be observed that with the increase in average acceleration values, energy consumption rate (ECR) generally increases. We can compute these 2 driver behaviour features in the following way :
\[\textit{Average Acceleration} = \frac{\sum_{i=1}^T a_i^+}{T}\]
\[\textit{Average Deceleration} = \frac{\sum_{i=1}^T a_i^-}{T}\]
where $a_i^+$ = Positive acceleration at time $i$, $a_i^+>0$, $a_i^-$ = Negative acceleration or deceleration at time $i$, $a_i^-<0$, and $T$ = Total time duration of the trip.

Average acceleration and deceleration are not enough to capture driver behaviour. Another trip feature that is useful to quantify driver behaviour is \textit{Relative Positive Acceleration (RPA)}. \cite{younes_analysis_2013} discussed the importance of RPA. RPA measures how much the driver accelerates in different speed values. 
\[
RPA = \frac{1}{d}\sum_{i=1}^T v_i\cdot a_i^+
\]
where $v_i$ is speed at time $i$, $a_i^+$ is the positive acceleration at time $i$, $a_i^+ > 0$, $T$ is the time duration of the trip and $d$ is the total trip distance. Aggressive drivers tend to accelerate more even at a higher speed, whereas a calm driver is not so much prone to accelerate at a higher speed. So, an aggressive driver will have a higher RPA value than a calm driver. Figure \ref{rpa_fig} illustrates that increase in RPA results in increase in energy consumption. 

\subsubsection{Environmental Factors}

\begin{figure}[h!]
    \subfigure[Energy Consumption Rate for different Positive Elevation Change values]{ \includegraphics[width=7cm]{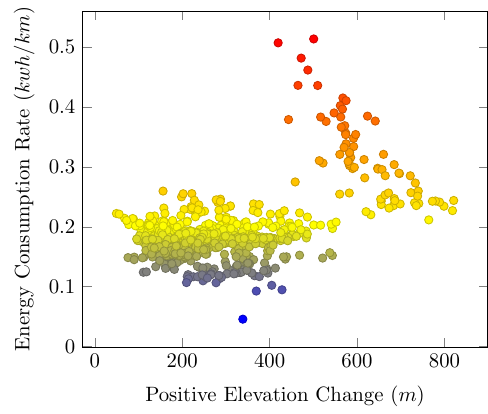} \label{pos_fig}}\hspace{5pt}
    \subfigure[Energy Consumption Rate for different Negative Elevation Change values]{\includegraphics[width=7cm]{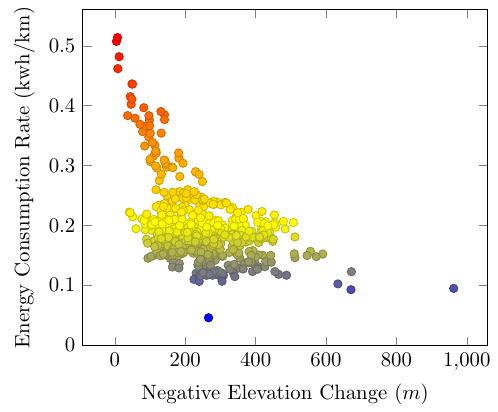} \label{neg_fig}}\\
    \begin{center}
    \subfigure[Energy Consumption Rate for different Speed and Temperature values]{\includegraphics[width=8cm]{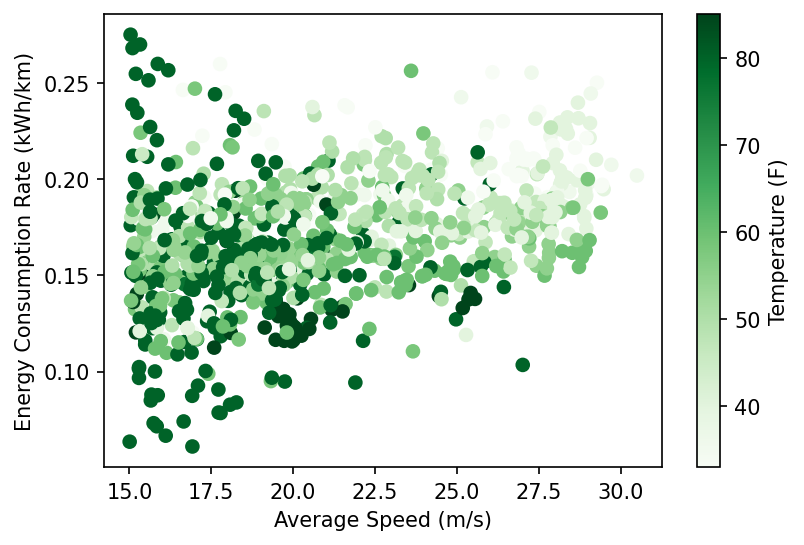} \label{temp_speed_fig}}
    \end{center}
    \caption{Energy Consumption Rate for different Environmental factor values}
    \label{env_analysis}
\end{figure}

Environmental Factors can influence the energy consumption of EVs heavily \citep{zhang2020energy}. Some of the most important environmental factors are :
\begin{itemize}
    \item \textit{Weather} :
    In low temperature, the ability of the battery to supply and receive current reduces \citep{younes_analysis_2013}. The energy consumption rate is generally higher during winter due to this reduction in battery efficiency. As a result this, Range of EV reduces in colder weather \citep{krogh2015analyzing}. We used temperature as an input factor to capture this effect. In Figure \ref{temp_speed_fig}, we have plotted energy consumption rate (ECR) for different speed and temperature values for the trips with average speed $\geq 15m/s$. It can be observed that ECR values are lower for trips with higher temperature. Though wind speed and wind direction also play a crucial role in energy consumption, it is generally difficult to precisely determine the speed and direction of wind. As ChargeCar dataset didn't record any information regarding wind we could not use wind speed as an input factor. Raining can also impact the EV energy consumption \citep{tannahill2016driver}. The rolling resistance on a wet road is higher than that on a dry road. This can potentially increase the energy consumption rate. As rain (or precipitation) data was not collected consistently for all the trips in ChargeCar dataset, we could not use rain as a factor in our model.
    \item \textit{Elevation} :
     Energy consumption increases while going uphill. While going downhill, the energy consumption will be lower. If the slope of the downhill is very steep, the battery current direction might reverse and a small fraction of the change in potential energy might get stored in the battery. This is called energy regeneration. As positive elevation change and negative elevation change affect energy consumption differently, we consider these 2 factors independently as mentioned in \cite{de_cauwer_energy_2015} and \cite{cauwer_2018}. Positive Elevation Change and Negative Elevation Change for a specific trip with time duration $T$ is defined as :
     \[\textit{Positive Elevation Change} = \sum_{i=1}^{T-1} max(h_{i+1}-h_i,0)\]
    \[\textit{Negative Elevation Change} = \sum_{i=1}^{T-1} min(h_{i+1}-h_i,0)\] 
    Where $h_j$ : elevation at time $j$ and $T$ : total time duration of the trip.
    In Figure \ref{pos_fig}, we have plotted energy consumption rates (ECR) and positive elevation change for different trips. ECR generally increases with increase in positive elevation change. ECR generally decreases with increase in negative elevation change, which can be seen in Figure \ref{neg_fig}. 
    \item \textit{Road Type} :
    Energy consumption also depends on the Road Type of the trip. In Highways, energy consumption rate is expected to be much lower compared to Urban Roads, as Urban Roads generally have more traffic congestion. Rolling resistance, which directly influences the energy consumption, also depends on road type. As road type and rolling resistance data were not present in the ChargeCar data, we did not use these features.
\end{itemize}

\subsection{Trip Characteristic Features}
\label{features_subsec}
As we have discussed in the previous section, there are various different factors that influence the energy consumption. It is not possible to record  all those factors, while collecting the data. Based on the availability of the features in ChargeCar dataset, we have considered the following input trip features for predicting energy consumption : 
\begin{enumerate}
    \item Average Speed ($m/s$)
    \item Standard Deviation of Speed ($m/s$)
    \item Trip Distance ($m$)
    \item Positive Elevation Change ($m$)
    \item Negative Elevation Change ($m$)
    \item Temperature ($^{\circ}F$)
    \item Relative Positive Acceleration ($m/s^2$)
    \item Average Acceleration ($m/s^2$)
    \item Average Deceleration ($m/s^2$)
\end{enumerate}

Table \ref{features_table} lists all the different types of input features used in this study and Table \ref{notations_table} contains all the notations used in this paper. 

\renewcommand{\arraystretch}{1.5}

\begin{table}[h!]
\tbl{Input Trip Characteristic Features}
{\scalebox{1.2}{
\begin{tabular}{lc}
\hline
\multicolumn{1}{c}{\textbf{Input Feature}} &  \textbf{Type}                          \\ \hline
Average Speed           &             \multirow{2}{*}{Vehicle Dynamics Factors}      \\ \cline{1-1}
Standard Deviation of Speed     &                                        \\ \hline
Trip Distance                &                                        \\ \hline
Positive Elevation Change       & \multirow{3}{*}{Environmental Factors} \\ \cline{1-1}
Negative Elevation Change     &                                        \\ \cline{1-1}
Temperature                 &                                        \\ \hline
Relative Positive Acceleration (RPA)             & \multirow{3}{*}{Driver Behaviour Factors}      \\ \cline{1-1}
Average Acceleration              &                                        \\ \cline{1-1}
Average Deceleration     &                                  \\ \hline
\end{tabular}}}
\label{features_table}
\end{table}

\begin{table}[h!]
\tbl{List of notations}
{\scalebox{1.2}{\begin{tabular}{ll}
\hline
\textbf{Notation} & \textbf{Description}                      \\ \hline
$V_{avg}$         & Average speed                             \\ \hline
$V_{std}$         & Standard deviation of speed               \\ \hline
$a_i^{+}$         & Positive acceleration at time $i$         \\ \hline
$a_i^{-}$         & Negative acceleration at time $i$         \\ \hline
RPA               & Relative positive acceleration            \\ \hline
$T$               & Time duration of a trip                   \\ \hline
$d$               & Trip distance                             \\ \hline
$p(.)$            & Probability distribution                  \\ \hline
$\mathcal{N}(.)$  & Gaussian distribution                     \\ \hline
$\mathcal{D}$     & Training data                             \\ \hline
$\mathbf{W}$      & Set of all weights in a neural network    \\ \hline
$M$               & Total number of NNs in an ensemble of NNs \\ \hline
\end{tabular}}}
\label{notations_table}
\end{table}

\section{Probabilistic Neural Network with Model  Uncertainty for Energy Consumption Estimation}
EV energy consumption is highly sensitive to numerous internal and external factors. Even a slight deviation in any of the contributing factors can abruptly change the EV energy consumption value. Because of this variability and uncertainty in energy consumption, estimated values from deterministic models \citep{zheng_hybrid_2016, qi_data-driven_2018, cauwer_2018} are not always reliable and accurate under real-world conditions. In order to address this challenge, we introduce a probabilistic neural network with weight uncertainty-based method for EV trip energy consumption prediction. 

To quantify uncertainty in energy consumption, the EV energy model has to predict a probability distribution of energy consumption instead of a point estimate, as shown in Figure \ref{prob_model}.
For a given \textit{Trip Feature Vector} $x$, the output of the Probabilistic Energy Consumption model will be a probability distribution $P(y|x;z)$, where $z$ is the set of parameters of the distribution.
\[
y|x \sim P(y|x;z)
\]

\input{Figures/prob}

Instead of estimating an arbitrary probability distribution, generally a specific known distribution is estimated. \cite{wu2015electric} showed that EV power consumption within a specific range of speed, acceleration and road inclination fits a Gaussian distribution. In our case, we considered Gaussian distribution for energy consumption. So, we don't have to estimate the whole distribution. We just have to predict the parameters of the Gaussian distribution, which are mean $\mu_{y|x}$ and standard deviation $\sigma_{y|x}$ in this case.
\begin{equation}
y|x \sim \mathcal{N}(\mu_{y|x},\sigma_{y|x})
\end{equation}

\begin{figure}[h!]
    \centering
    \includegraphics{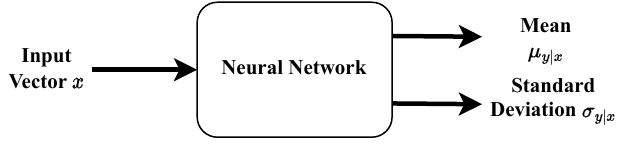}
    \caption{Neural Network with probabilistic output}
    \label{nn_img1}
\end{figure}

If the posterior distribution $P(y|x;z)$ is considered to be a Gaussian distribution, the objective is to implement and train a neural network (NN) to estimate only the distribution parameters : mean $\mu_{y|x}$ and standard deviation $\sigma_{y|x}$, as illustrated by Figure \ref{nn_img1}.

If the training data is $\mathcal{D} = \{x_i,y_i\}_{i=1}^N$, the predicted outputs for training data point $x_i$ are $\mu_i$ and $\sigma_i$ and $\mathbf{W}$ is the set of weights in the NN, the likelihood distribution is $p(\mathcal{D}|\mathbf{W}) = \prod_{i}^{N} \mathcal{N}(y_i|
\mu_i,\sigma_i)$. Negative log likelihood loss function is used as the objective function to train probabilistic neural networks. Negative log likelihood loss function is computed as :
\begin{equation}
\begin{split}
        \mathcal{L}(\mathbf{W}) & = - \log        
        \prod_{i=1}^{N} \mathcal{N}(y_i \mid \mu_i,\sigma_i)\\
        & = \sum_{i=1}^{N} (\log \sigma_i+\frac{(y_i-\mu_i)^2}{2\sigma_i^2}) + C 
\end{split}
\end{equation}
where $C$ is a constant.
The optimization problem is :
\begin{equation}
\begin{split}
        \mathbf{W}^* & = \argmin_{\mathbf{W}} \mathcal{L}(\mathbf{W})\\
        & = \argmin_{\mathbf{W}} \sum_{i=1}^{N} (\log \sigma_i+\frac{(y_i-\mu_i)^2}{2\sigma_i^2})  
\end{split}
\end{equation}

\subsection{Model Uncertainty or Weight Uncertainty in Neural Network}
In deterministic machine learning models or neural networks, the model parameters are considered point estimates. In contrast, Bayesian probabilistic machine learning methods treat model parameters as random variables. So, Bayesian methods learn probability distributions of model parameters. This is called \textit{Model  Uncertainty} \citep{KIUREGHIAN2009105} in machine learning models or weight uncertainty in Neural Networks. In neural networks, model uncertainty or weight uncertainty can be incorporated by considering probability distributions for weights. \cite{hinton1993keeping} and \cite{blundell2015weight} have shown that weight uncertainty helps NNs avoid overfitting and makes NNs much more robust to noise and outliers. As the EV energy consumption problem also deals with numerous uncertain factors and the problem of overfitting, noise and outliers, we propose a probabilistic neural network with weight uncertainty model for this task.

If the training data is $\mathcal{D} = \{x_i,y_i\}_{i=1}^N$, $\mathbf{W}$ is the set of weights in the NN and $p(\mathbf{W})$ is the prior distribution for the NN weights, the posterior distribution for the weights is given by :
\begin{equation}
p(\mathbf{W}|\mathcal{D}) = \frac{p(\mathcal{D}|\mathbf{W})p(\mathbf{W})}{\int_{\mathbf{W'}} p(\mathcal{D}|\mathbf{W'})p(\mathbf{W'})d\mathbf{W'}}
\end{equation}

The posterior distribution $p(\mathbf{W}|\mathcal{D})$ is generally intractable and highly expensive to compute. \cite{hinton1993keeping} proposed variational inference methods to approximate the intractable distribution $p(\mathbf{W}|\mathcal{D})$ using a tractable distribution $q(\mathbf{W}|\phi)$, where $\phi$ is the set of distribution parameters.

To incorporate weight uncertainty in neural network using variational inference, \textit{Evidence Lower Bound (ELBO)} loss function is used to quantify the error in prediction. In neural networks with weight uncertainty, the objective is to learn a posterior distribution $q(\mathbf{W}|\phi)$ over the set of weights $\mathbf{W}$, where $\phi$ is the set of parameters of the probability distribution, given a prior distribution $p(\mathbf{W})$ and likelihood distribution $p(\mathcal{D}|\mathbf{W})$. ELBO loss function $J(\mathbf{W})$ is computed as :
\begin{equation}
\begin{split}
J(\mathbf{W}) & = \overbrace{-\mathbbm{E}_{q(\mathbf{W}|\phi)}[p(\mathcal{D}|\mathbf{W})]}^\text{Negative Log Likelihood} + \overbrace{KL(q(\mathbf{W}|\phi)||p(\mathbf{W}))}^\text{KL Divergence of prior and posterior}\\
& \approx  \frac{1}{N}\sum_{i=1}^N [ - p(\mathcal{D}|\mathbf{W}^i) + \log  q(\mathbf{W}^i|\phi) - \log p(\mathbf{W}^i)]
\end{split}
\end{equation}

ELBO loss function consists of the Negative Log Likelihood and Kullback–Leibler divergence (KL divergence), which are approximated by sampling $N$ different values of weights $\{\mathbf{W}^i\}_{i=1}^N$. In our case, we considered Gaussian distributions for both $q(\mathbf{W}|\phi)$ and $p(\mathbf{W})$, so $\phi = \{(\mu^w,\sigma^w)|\forall w \in \mathbf{W}\}$. Backpropagation algorithm is used to train the neural network and learn the posterior distribution $q(\mathbf{W}|\phi)$ for the set of weights.

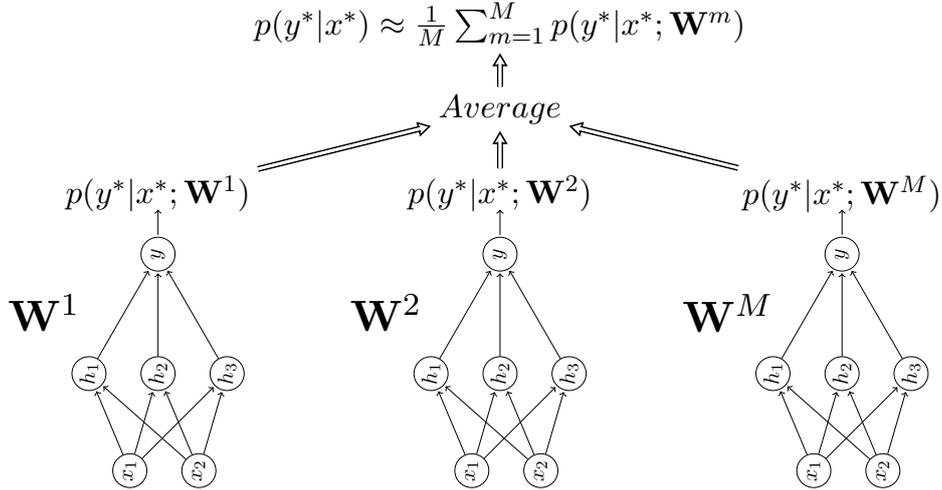
\begin{figure*}[htbp]
\centering
\scalebox{0.75}{
\begin{tikzpicture}[
roundnode/.style={circle, draw=green!60, fill=green!5, very thick, minimum size=7mm},
squarednode/.style={rectangle, draw=red!60, fill=red!5, very thick, minimum size=2cm},
newnode/.style={rectangle}
]
\tikzstyle{vecArrow} = [thick, decoration={markings,mark=at position
   1 with {\arrow[thick]{open triangle 60}}},
   double distance=1.4pt, shorten >= 5.5pt,
   preaction = {decorate},
   postaction = {draw,line width=1.4pt, white,shorten >= 4.5pt}]
\tikzstyle{myarrows}=[line width=0.8mm,draw=gray,-triangle 45,postaction={draw,line width=2mm, shorten >=4mm, -}]

\node [align=center,font=\bfseries] at (0,7) (1) {\scalebox{1.5}{$Average$}};
\node [align=center,font=\bfseries,rotate=90] at (0,3) (2) {\begin{tikzpicture}[shorten >=1pt,->, draw=black!100,
        node distance = 6mm and 15mm,
          start chain = going below,
every pin edge/.style = {<-,shorten <=1pt},
        neuron/.style = {circle, draw=black, fill=#1,   
                         minimum size=17pt, inner sep=0pt,
                         on chain},
         annot/.style = {text width=4em, align=center}
                        ]
\foreach \i in {1,...,2}
    \node[neuron=white!50
          ] (I-\i)    {$x_{\i}$};
    \node[neuron=white!50,
          above right=5mm and 15mm of I-1.center] (H-1)     {$h_{1}$};
\foreach \i [count=\j from 1] in {2,...,3}
    \node[neuron=white!50,
          below=of H-\j]      (H-\i)    {$h_{\i}$};
    \node[neuron=white!50,
        pin= {[pin edge=->]0: },
          right=of H-2]  (O-1)   {$y$};
    \foreach \i in {1,...,2}
        \foreach \j in {1,...,3}
{
    \path (I-\i) edge (H-\j)
    \ifnum\i<2                  
          (H-\j) edge (O-\i)
    \fi;
}
\end{tikzpicture}};
\node[align=center] at (-2,3.4) (-2) {\scalebox{2}{$\mathbf{W}^2$}};
\node[align=center] at (0,5.5) (3) {\scalebox{1.5}{$p(y^*|x^*;\mathbf{W}^2)$}};
\node [align=center,font=\bfseries,rotate=90] at (-6,3) (4) {\begin{tikzpicture}[shorten >=1pt,->, draw=black!100,
        node distance = 6mm and 15mm,
          start chain = going below,
every pin edge/.style = {<-,shorten <=1pt},
        neuron/.style = {circle, draw=black, fill=#1,   
                         minimum size=17pt, inner sep=0pt,
                         on chain},
         annot/.style = {text width=4em, align=center}
                        ]
\foreach \i in {1,...,2}
    \node[neuron=white!50
          ] (I-\i)    {$x_{\i}$};
    \node[neuron=white!50,
          above right=5mm and 15mm of I-1.center] (H-1)     {$h_{1}$};
\foreach \i [count=\j from 1] in {2,...,3}
    \node[neuron=white!50,
          below=of H-\j]      (H-\i)    {$h_{\i}$};
    \node[neuron=white!50,
        pin= {[pin edge=->]0: },
          right=of H-2]  (O-1)   {$y$};
    \foreach \i in {1,...,2}
        \foreach \j in {1,...,3}
{
    \path (I-\i) edge (H-\j)
    \ifnum\i<2                  
          (H-\j) edge (O-\i)
    \fi;
}
\end{tikzpicture}};
\node[align=center] at (-8,3.4) (-2) {\scalebox{2}{$\mathbf{W}^1$}};
\node[align=center] at (-6,5.5) (5) {\scalebox{1.5}{$p(y^*|x^*;\mathbf{W}^1)$}};
\node [align=center,font=\bfseries,rotate=90] at (6,3) (6) {\begin{tikzpicture}[shorten >=1pt,->, draw=black!100,
        node distance = 6mm and 15mm,
          start chain = going below,
every pin edge/.style = {<-,shorten <=1pt},
        neuron/.style = {circle, draw=black, fill=#1,   
                         minimum size=17pt, inner sep=0pt,
                         on chain},
         annot/.style = {text width=4em, align=center}
                        ]
\foreach \i in {1,...,2}
    \node[neuron=white!50
          ] (I-\i)    {$x_{\i}$};
    \node[neuron=white!50,
          above right=5mm and 15mm of I-1.center] (H-1)     {$h_{1}$};
\foreach \i [count=\j from 1] in {2,...,3}
    \node[neuron=white!50,
          below=of H-\j]      (H-\i)    {$h_{\i}$};
    \node[neuron=white!50,
        pin= {[pin edge=->]0: },
          right=of H-2]  (O-1)   {$y$};
    \foreach \i in {1,...,2}
        \foreach \j in {1,...,3}
{
    \path (I-\i) edge (H-\j)
    \ifnum\i<2                  
          (H-\j) edge (O-\i)
    \fi;
}
\end{tikzpicture}};
\node[align=center] at (4,3.4) (-2) {\scalebox{2}{$\mathbf{W}^M$}};
\node[align=center] at (6,5.5) (7) {\scalebox{1.5}{$p(y^*|x^*;\mathbf{W}^M)$}};
\node [align=center,font=\bfseries] at (0,8.5) (8) {\scalebox{1.5}{$p(y^*|x^*) \approx  \frac{1}{M} \sum_{m=1}^M p(y^*|x^*;\mathbf{W}^m)$}};

\draw[vecArrow] (1) to (8);
\draw[vecArrow] (3) to (1);  
\draw[vecArrow] (5) to (1);
\draw[vecArrow] (7) to (1);
\end{tikzpicture}}
\caption{Predictive Posterior Distribution prediction using Monte Carlo sampling and Ensemble of Neural Networks}
\label{ensemble_nn}
\end{figure*}

\subsection{Predictive Posterior Distribution using Monte Carlo Sampling and Ensemble of NNs}
As the weight values are generated from the posterior probability distributions in NN with weight uncertainty, the output of the NN is also uncertain. If we make $M$ predictions with the NN on the same input data point $x^*$, we might get a different output on every run. Because of this weight uncertainty, we have an ensemble of infinite number of neural networks. As it is not feasible to get the outputs of infinite number of neural networks, we have used Monte Carlo sampling to get the final output. The predictive posterior distribution is calculated as :
\begin{equation}
\begin{split}
p(y^* \mid x^*) & = \int_{\mathbf{W}} p(y^* \mid x^*;\mathbf{W})p(\mathbf{W} \mid \mathcal{D}) d\mathbf{W} \\
& = \int_{\mathbf{W}} p(y^* \mid x^*;\mathbf{W})q(\mathbf{W} \mid \phi) d\mathbf{W} \\
& = \mathbb{E}_{\mathbf{W} \sim q(\mathbf{W}|\phi)}[p(y^*|x^*;\mathbf{W})]\\
& \approx \frac{1}{M}\sum_{m=1}^{M}p(y^*|x^*;\mathbf{W}^m)
\end{split}
\end{equation}

We approximate the outputs of infinite number of neural networks with an ensemble of finite number of neural networks, whose weights are sampled using Monte Carlo method. $M$ different weight values are sampled from the posterior distribution $q(\mathbf{W}|\phi)$ : $\mathbf{W}^1,\mathbf{W}^2,...,\mathbf{W}^M \sim q(\mathbf{W}|\phi)$. If $y^*_m$ is the output with weight value $\mathbf{W}^m$ and $y^*_m \sim \mathcal{N}(\mu^*_m,\sigma^*_m)$, the final output $y^*$ is the average of the $M$ outputs :
\begin{equation}
\begin{split}
y^*  = \frac{1}{M}\sum_{m=1}^{M}y^*_m\\
y^* \sim \mathcal{N}(\mu^*,\sigma^*)
\end{split}
\end{equation}
where $\mu^* = \frac{1}{M}\sum_m{\mu_m^*}$, $\sigma^{*2} = \frac{1}{M^2}\sum_m{\sigma^{*2}}$. So, the predictive posterior distribution is $p(y^*|x^*)$, where $y^*$ is a Gaussian random variable with mean $\mu^*$ and variance $\sigma^{*2}$. The predictive posterior distribution computation using Monte Carlo sampling is shown in Figure \ref{ensemble_nn}.

As we are able to create an ensemble of neural networks with the use of weight uncertainty, our method improves the accuracy of energy consumption prediction and robustness of the model against noise and outliers under real-world conditions compared to any single energy estimation model.

\begin{figure}[h!]
    \centering
    \includegraphics[width=\linewidth]{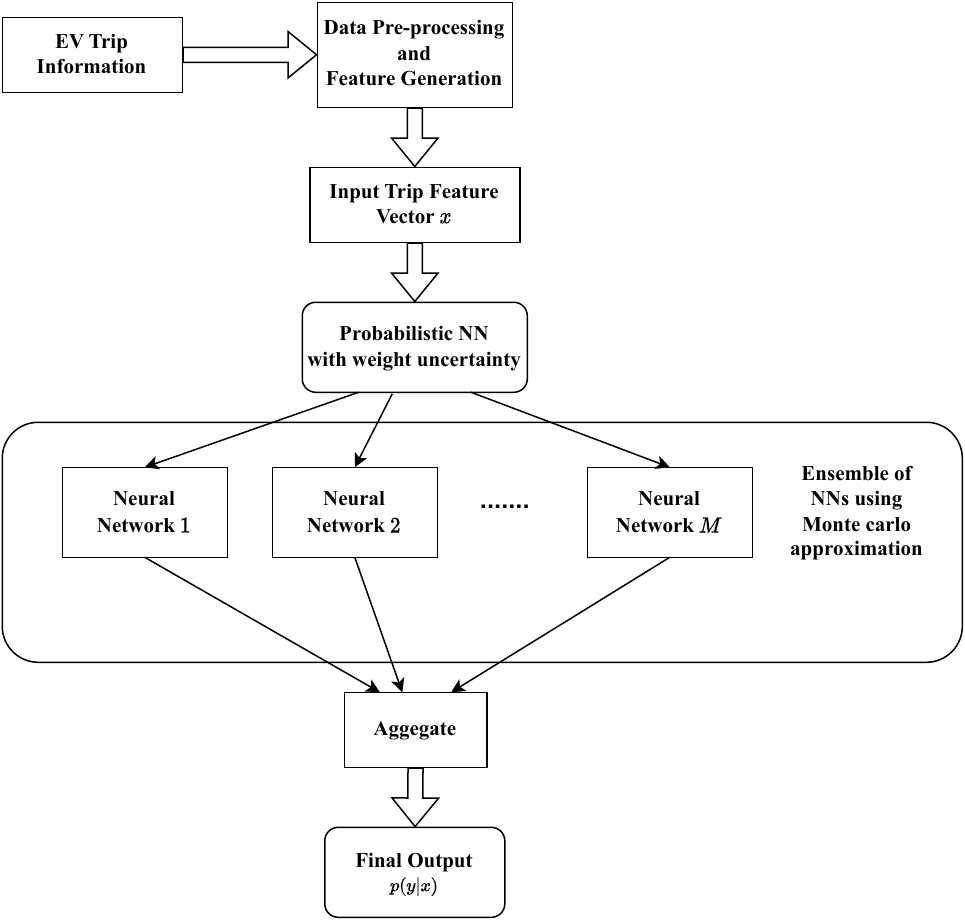}
    \caption{Conceptual framework of our proposed methodology}
    \label{method_fig}
\end{figure}

The conceptual framework of our proposed method is illustrated in Figure \ref{method_fig}. At first, EV trip information containing vehicle speed, acceleration, weather and other data is pre-processed and trip input features are generated, as detailed in Section \ref{features_subsec}. The generated input trip feature vector $x$ goes as input to the probabilistic NN with weight uncertainty. An ensemble of $M$ number of NNs is created using Monte Carlo approximation. Finally, the output from each NN is aggregated as shown in Figure \ref{ensemble_nn} and the final posterior distribution for energy consumption $p(y|x)$ is computed.   

\section{Experimental Results}

\begin{table}[htbp]
\centering
\tbl{Descriptive Statistics of input and output variables of the data}
{\scalebox{1.1}{\begin{tabular}{lllll}
\hline
                                & \textbf{Variables}                        & \textbf{Max} & \textbf{Min} & \textbf{Average} \\ \hline
\multirow{9}{*}{\textbf{Input Features}} & Average Speed ($m/s$)                    & 30.4         & 1.5          & 11.6             \\ \cline{2-5} 
                                & Standard Deviation of Speed ($m/s$)      & 14.1         & 0.4          & 6.3              \\ \cline{2-5} 
                                & Trip Distance ($m$)                      & 6752.3       & 0.08         & 1124.0           \\ \cline{2-5} 
                                & Positive Elevation Change ($m$)          & 821.1        & 2.4          & 161.0            \\ \cline{2-5} 
                                & Negative Elevation Change ($m$)          & 0            & -962.8       & -150.7           \\ \cline{2-5} 
                                & Temperature ($^{\circ}F$)                        & 85.0         & 33.0         & 62.4             \\ \cline{2-5} 
                                & Relative Positive Acceleration ($m/s^2$) & 0.25         & 0.01         & 0.09             \\ \cline{2-5} 
                                & Average Acceleration ($m/s^2$)           & 0.41         & 0.028        & 0.18             \\ \cline{2-5} 
                                & Average Deceleration ($m/s^2$)           & -0.024       & -0.387       & -0.187           \\ \hline
\textbf{Output}                  & Energy ($kWh$)                             & 11.2         & -1.24        & 1.84             \\ \hline
\end{tabular}}}
\label{stats}
\end{table}

\subsection{Data Preprocessing}
ChargeCar EV trip dataset \citep{chargecar_data} has to be preprocessed in order to use it to train our proposed model. As the number of trips in the dataset is quite less to train machine learning models, we divided the trips with 3 levels of randomness as shown in \citep{zheng_hybrid_2016}.  A trip is randomly picked from the dataset. Then we take a random starting point in the trip and cut out a random length micro-trip \citep{zheng_hybrid_2016} from the complete trip.  From the 50 EV trips, we generated 5000 micro-trips. We removed the micro-trips with very low energy consumption (consumption $<$ 0.3kWh) or energy regeneration (regeneration $<$ 0.3 kWh). After this filtering, we got a total of 3916 micro-trips. We used 90\% of those micro-trips for training and rest of micro-trips are used for testing. We have presented the descriptive statistics (i.e. minimum, maximum and average) of different input variables and output variable (i.e. Energy consumption) in Table \ref{stats}.

To train the NN smoothly, we also normalized the training data. Min-max scaling is used to normalize the data. For a given trip feature vector $x_i$, the mini-max scaling is given by :
\begin{equation}
x_i^{scaled} = \frac{x_i-min(x_i)}{max(x_i)-min(x_i)}
\end{equation}

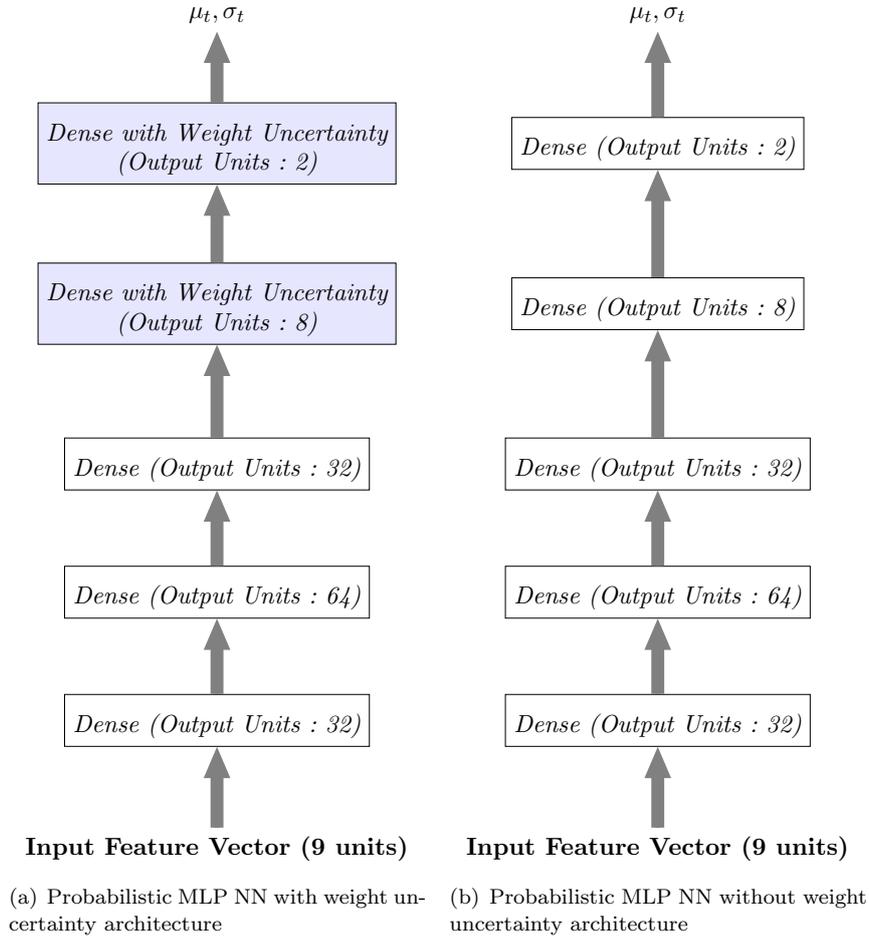
\begin{figure}[h!]
\begin{center}
    \subfigure[Probabilistic MLP NN with weight uncertainty architecture]{\scalebox{0.85}{
    \begin{tikzpicture}[
    roundnode/.style={circle, draw=green!60, fill=green!5, very thick, minimum size=7mm},
    squarednode/.style={rectangle, draw=red!60, fill=red!5, very thick, minimum size=2cm},
    newnode/.style={rectangle}
    ]
    \tikzstyle{vecArrow} = [thick, decoration={markings,mark=at position
       1 with {\arrow[thick]{open triangle 60}}},
       double distance=1.4pt, shorten >= 5.5pt,
       preaction = {decorate},
       postaction = {draw,line width=1.4pt, white,shorten >= 4.5pt}]
    \tikzstyle{myarrows}=[line width=0.8mm,draw=gray,-triangle 45,postaction={draw,line width=2mm, shorten >=4mm, -}]
    
    \node [align=center,font=\bfseries] at (0,-3) (0) {Input Feature Vector (9 units)};
    \node [draw,align=center,font=\itshape] at (0,-1.0) (1) {\\Dense (Output Units : 32)};
    \node [draw,align=center,font=\itshape] at (0,1) (2) {\\Dense (Output Units : 64)};
    \node [draw,align=center,font=\itshape] at (0,3) (3) {\\Dense (Output Units : 32)};
    \node [draw,fill=blue!10,align=center,font=\itshape] at (0,5.5) (4) {\\Dense with Weight Uncertainty\\(Output Units : 8)};
    \node [draw,fill=blue!10,align=center,font=\itshape] at (0,8.0) (5) {\\Dense with Weight Uncertainty\\(Output Units : 2)};
    \node [align=center,font=\itshape] at (0,10.0) (6) {$\mu_{t},\sigma_t$};
    \draw[myarrows] (0) to (1);
    \draw[myarrows] (1) to (2);
    \draw[myarrows] (2) to (3);  
    \draw[myarrows] (3) to (4); 
    \draw[myarrows] (4) to (5);
    \draw[myarrows] (5) to (6);
    \end{tikzpicture}} \label{nn_prob1} }\hspace{5pt}
    \subfigure[Probabilistic MLP NN without weight uncertainty architecture]{\scalebox{0.85}{
\begin{tikzpicture}[
roundnode/.style={circle, draw=green!60, fill=green!5, very thick, minimum size=7mm},
squarednode/.style={rectangle, draw=red!60, fill=red!5, very thick, minimum size=2cm},
newnode/.style={rectangle}
]
\tikzstyle{vecArrow} = [thick, decoration={markings,mark=at position
   1 with {\arrow[thick]{open triangle 60}}},
   double distance=1.4pt, shorten >= 5.5pt,
   preaction = {decorate},
   postaction = {draw,line width=1.4pt, white,shorten >= 4.5pt}]
\tikzstyle{myarrows}=[line width=0.8mm,draw=gray,-triangle 45,postaction={draw,line width=2mm, shorten >=4mm, -}]

\node [align=center,font=\bfseries] at (0,-3) (0) {Input Feature Vector (9 units)};
\node [draw,align=center,font=\itshape] at (0,-1.0) (1) {\\Dense (Output Units : 32)};
\node [draw,align=center,font=\itshape] at (0,1) (2) {\\Dense (Output Units : 64)};
\node [draw,align=center,font=\itshape] at (0,3) (3) {\\Dense (Output Units : 32)};
\node [draw,align=center,font=\itshape] at (0,5.5) (4) {\\Dense (Output Units : 8)};
\node [draw,align=center,font=\itshape] at (0,8.0) (5) {\\Dense (Output Units : 2)};
\node [align=center,font=\itshape] at (0,10.0) (6) {$\mu_{t},\sigma_t$};
\draw[myarrows] (0) to (1);
\draw[myarrows] (1) to (2);
\draw[myarrows] (2) to (3);  
\draw[myarrows] (3) to (4); 
\draw[myarrows] (4) to (5);
\draw[myarrows] (5) to (6);
\end{tikzpicture}} \label{nn_prob2}}
\caption{Probabilistic MLP NN Architectures}
    \label{fig:my_label}
\end{center}
\end{figure}

\subsection{Implementation and Training of the Models}

We have implemented 3 different Multilayer Perceptron (MLP) Neural Network (NN) models: 
\begin{enumerate}
    \item \textit{Probabilistic MLP NN with Weight Uncertainty} : This model has 4 hidden layers as shown in the Figure \ref{nn_prob1} and only the last two layers have weight uncertainty. Numbers of units in 4 hidden layers are : 32, 64, 32, 8. The output layer consists of 2 output units. We considered weight uncertainty in the last 2 layers as incorporating uncertainty in all layers increased the model complexity but did not show any improvement in results. 
    \item \textit{Probabilistic MLP NN without Weight Uncertainty} : The architecture of this model is very  similar to the previous model. Figure \ref{nn_prob2} depicts the architecture for the probabilistic MLP NN without weight uncertainty. The number of layers and number of units in each layers are same as the previous model. The only difference is that it doesn't have any weight certainty.
    \item \textit{Deterministic MLP NN} : The deterministic MLP NN contains 4 layers with number of hidden units : 32, 64, 32, 8 and it has only 1 output unit as it is a deterministic model.
\end{enumerate}

Each of these 3 models have been implemented and trained in 2 different scenarios : \textit{With Driver Behaviour Features} and \textit{\textit{Without Driver Behaviour Features}}. The implementation of the models and experiments were carried out in Python programming language. The neural  networks are implemented using Tensorflow and Tensorflow Probability \citep{tensorflow2015-whitepaper,dillon2017tensorflow} and trained using Backpropagation algorithm with Adam optimizer with learning rate = 0.05. The neural networks are trained for 400 epochs. To compute the final posterior predictive distribution using Monte Carlo sampling, we have taken 10 random samples.

\subsection{Evaluation Metrics}
We used 2 evaluation metrics : Root Mean Squared Error (\textit{RMSE}) and Mean Absolute Percentage Error (\textit{MAPE}).
\[
RMSE(y,\hat{y}) = \sqrt{\frac{1}{N}\sum_{i=1}^{N}(y_i-\hat{y_i})^2}
\]
\[\small
MAPE(y,\hat{y}) = \frac{1}{N}\sum_{i=1}^{N}\mid \frac{ y_i-\hat{y_i}}{y_i}\mid * 100
\]
where $y_i$ is actual energy value for $i^{th}$ sample and $\hat{y_i}$ is the predicted energy value for $i^{th}$ sample. 

\renewcommand{\arraystretch}{1.5}

\begin{table}[htbp]
\centering
\tbl{RMSE and MAPE Error of Probabilistic and Deterministic MLP NN Models with and without Driver Behaviour}
{\scalebox{0.9}{\begin{tabular}{clcclcc}
\hline
\multicolumn{2}{c}{\multirow{2}{*}{\textbf{Method}}}               & \multicolumn{2}{c}{\textbf{With Driver Behaviour}} &  & \multicolumn{2}{c}{\textbf{Without Driver Behaviour}} \\ \cline{3-4} \cline{6-7} 
\multicolumn{2}{c}{}                                               & \textbf{MAPE(\%)}       & \textbf{RMSE(kWh)}       &  & \textbf{MAPE(\%)}         & \textbf{RMSE(kWh)}        \\ \hline
\multicolumn{2}{c}{Probabilistic MLP NN with Weight Uncertainty}   & \textbf{9.3}            & \textbf{0.20}            &  & \textbf{12.9}             & \textbf{0.24}             \\ \hline
\multicolumn{2}{c}{Probabilistic MLP NN without Weight Uncertainty} & 11.6                    & 0.23                     &  & 15.3                      & 0.33                      \\ \hline
\multicolumn{2}{c}{Deterministic MLP NN}                           & 14.2                    & 0.29                     &  & 20.2                      & 0.35                      \\ \hline
\end{tabular}}}
\label{comp1}
\end{table}

\renewcommand{\arraystretch}{1.5}

\begin{table}[h!]
\centering
\tbl{Comparing the results of our model with other EV energy consumption estimation methods}
{\scalebox{1.2}{\begin{tabular}{ll}
\hline
\textbf{Method}             & \textbf{MAPE(\%)} \\ \hline
Probabilistic MLP NN with Weight Uncertainty (\textbf{Our Approach}) & \textbf{9.3}     \\ \hline
M-GPR \citep{jiang2023trip} & 11.0 \\ \hline
Stochastic RF \citep{li2021prediction} & 11.8 \\ \hline
XGBoost \citep{zhang2020energy}  & 12.6 \\ \hline
NN-MLR \citep{cauwer_2018} & 14 \\ \hline
RT-SOM \citep{zheng_hybrid_2016}         & 25       \\ \hline
Regression \citep{de_cauwer_energy_2015}  & 25 \\ \hline
MLP \citep{shankar2013method} & 28 \\ \hline
Probabilistic DNN \citep{petkevicius_probabilistic_2021} & 52       \\ \hline
\end{tabular}}}

\label{comp2}
\end{table}

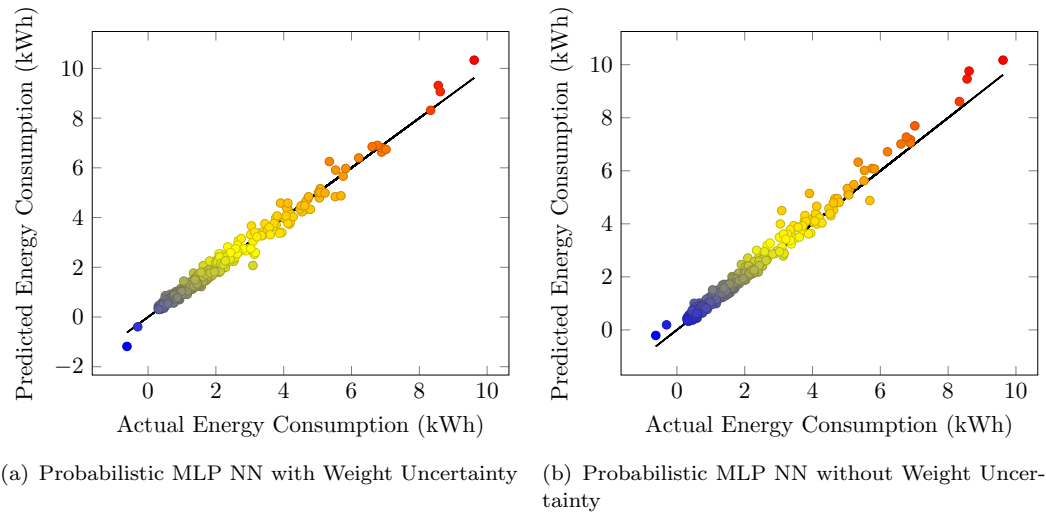
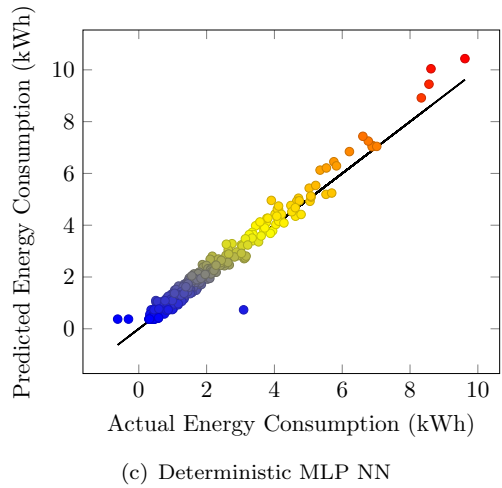
\begin{figure}[h!]
    \subfigure[Probabilistic MLP NN with Weight Uncertainty]{\scalebox{0.8}{
    \begin{tikzpicture}
    \begin{axis}[ylabel near ticks,xlabel=Actual Energy Consumption (kWh),ylabel=Predicted Energy Consumption (kWh)]
        \addplot[
                scatter, only marks]
         table[x expr=\thisrow{c1}/3600.0,y expr=\thisrow{c2}/3600.0, col sep=comma]
            {Data/test_pred1.txt};
        \addplot[thick,ticks=none] table[x expr=\thisrow{c1}/3600.0,y expr=\thisrow{c1}/3600.0,col sep=comma] {Data/test_pred1.txt};
    \end{axis}
    \end{tikzpicture}} \label{test_pred_fig1} }\hspace{5pt}
    \subfigure[Probabilistic MLP NN without Weight Uncertainty]{\scalebox{0.8}{\begin{tikzpicture}
    \begin{axis}[ylabel near ticks,xlabel=Actual Energy Consumption (kWh),ylabel=Predicted Energy Consumption (kWh)]
        \addplot[
                scatter, only marks]
         table[x expr=\thisrow{c1}/3600.0,y expr=\thisrow{c2}/3600.0, col sep=comma]
            {Data/test_pred2.txt};
        \addplot[thick,ticks=none] table[x expr=\thisrow{c1}/3600.0,y expr=\thisrow{c1}/3600.0,col sep=comma] {Data/test_pred2.txt};
    \end{axis}
    \end{tikzpicture} \label{test_pred_fig2} }}\\
    \begin{center}
    \subfigure[Deterministic MLP NN]{\scalebox{0.8}{\begin{tikzpicture}
    \begin{axis}[ylabel near ticks,xlabel=Actual Energy Consumption (kWh),ylabel=Predicted Energy Consumption (kWh)]
        \addplot[
                scatter, only marks]
         table[x expr=\thisrow{c1}/3600.0,y expr=\thisrow{c2}*1000.0/3600.0, col sep=comma]
            {Data/test_pred3.txt};
        \addplot[thick,ticks=none] table[x expr=\thisrow{c1}/3600.0,y expr=\thisrow{c1}/3600.0,col sep=comma] {Data/test_pred3.txt};
    \end{axis}
    \end{tikzpicture} \label{test_pred_fig3} }}
    \end{center}
    \caption{Actual and Predicted Trip Energy Consumption Comparison for 3 different methods}
    \label{test_pred_fig}
\end{figure}

\subsection{Results}

We have presented the results of the 3 models that we implemented in Table \ref{comp1}. We observe that models using driver behaviour are performing much better than models without it. It can be also seen in Table \ref{comp1} that Probabilistic MLP NN with weight uncertainty has outperformed all other models in both MAPE and RMSE error. Probabilistic MLP NN with weight uncertainty achieves a MAPE error of 9.3\%, while the MAPE error of probabilistic MLP NN without weight uncertainty is 11.6\%. Deterministic MLP NN is the worst performing model with a MAPE error of 14.2\%. RMSE error of probabilistic MLP NN with weight uncertainty and driven behaviour is 0.20 kWh.

In Table \ref{comp2}, we have compared our model with other EV energy consumption models. It can be observed that our model performed better than all other EV energy consumption methods. The second most accurate EV energy consumption according to Table \ref{comp2} is Markov-based Gaussian process regression \citep{jiang2023trip} with MAPE error of 11.0\%. Our proposed model is much more accurate than the Probabilistic DNN \citep{petkevicius_probabilistic_2021}, which had MAPE error of 52\% and their model did not consider model uncertainty and driver behaviour factors. 

From Table \ref{comp1} and \ref{comp2}, we can conclude that the use of Driver Behaviour features (RPA, Average Acceleration, Average Deceleration) and Environmental features (Temperature,Positive and Negative Elevation Change) can significantly improve the accuracy of EV energy consumption estimation.

As EV trip energy estimation is highly sensitive to various uncertain internal, external and driver-related factors, 
capturing the uncertainty in the input parameters and output is essential to create a more accurate model. The weight uncertainty helps our proposed probabilistic NN to capture energy consumption better than other methods. The weight uncertainty also enables our model to overcome the problem of overfitting \citep{hinton1993keeping,graves2011practical}, which makes our model much more accurate than others. By incorporating weight uncertainty, we have been able to create an ensemble of neural networks, which gives better results than probabilistic neural networks without weight uncertainty and deterministic neural networks, as shown in Table \ref{comp1}. 

Figure \ref{test_pred_fig} shows the actual and predicted trip energy consumption values for every test micro-trip using 3 different methods. From Figure \ref{test_pred_fig1}, we observe that energy consumption values predicted by our probabilistic model with weight uncertainty are very close to the actual values given in the test data. Figure \ref{test_pred_fig2} and Figure \ref{test_pred_fig3} have shown the actual and predicted energy consumption values for Probabilistic NN without weight uncertainty and Deterministic MLP NN respectively. From these 3 plots in Figure \ref{test_pred_fig}, we can clearly conclude that Probabilistic NN with uncertainty is much more accurate than other 2 methods.

\begin{table}[h!]
\centering
\caption{Training Times for Probabilistic and Determistic MLP NN models}
\begin{tabular}{ll}
\hline
\textbf{Method}                                          & \textbf{Training Time (s)} \\ \hline
Probabilistic MLP NN with Weight Uncertainty    & 189.2             \\ \hline
Probabilistic MLP NN without Weight Uncertainty & 150.1             \\ \hline
Deterministic MLP NN                            & 167.5             \\ \hline
\end{tabular}
\label{time_comp}
\end{table}

In Table \ref{time_comp}, the training times for the probabilistic and deterministic mlp nn models are reported. The probabilistic mlp nn with weight uncertainty took 189.2 seconds to train for 400 epoch. This shows that the training time for our probabilistic model is reasonably short.

\pgfplotsset{
    legend entry/.initial=,
    every axis plot post/.code={%
        \pgfkeysgetvalue{/pgfplots/legend entry}\tempValue
        \ifx\tempValue\empty
            \pgfkeysalso{/pgfplots/forget plot}%
        \else
            \expandafter\addlegendentry\expandafter{\tempValue}%
        \fi
    },
}

\begin{figure}[h!]
    \subfigure[ECR for different $V_{avg}$ and $V_{std}$]{\scalebox{0.75}{
    \begin{tikzpicture}
        \begin{axis}[x dir=reverse,ylabel = $V_{std}(m/s)$,
            xlabel = $V_{avg}(m/s)$,zlabel = ECR($kwh/km$),width=7cm]
        \addplot3 [surf, mesh/rows=20, shader=interp]
          table[x=c1, y=c2, z=c3, col sep=comma] {Data/VDM1.tex};
        \end{axis}
    \end{tikzpicture}} \label{ecr_vdm} }\hspace{5pt}
    \subfigure[ECR for different temperatures]{\scalebox{0.75}{
    \begin{tikzpicture}
    \begin{axis}[ylabel=ECR($kwh/km$),xlabel=Temperature(${}^\circ F$),width=8cm]
    \addplot [thick, ticks=none] table [x=c1, y=c2, col sep=comma] {Data/temp.txt};
    \addplot [name path=upper,draw=none] table [x=c1, y expr= \thisrow{c2}+3*\thisrow{c3}-3*\thisrow{c2}, col sep=comma] {Data/temp.txt};
    \addplot [name path=lower,draw=none] table [x=c1, y expr = \thisrow{c2}+3*\thisrow{c4}-3*\thisrow{c2}, col sep=comma] {Data/temp.txt};
    \addplot [fill=blue!20,legend entry=\text{95\% Confidence Interval}] fill between[of=upper and lower];
    \end{axis}
    \end{tikzpicture}} \label{ecr_temp} }\\
    \subfigure[ECR for different Driver Behaviour RPA]{\scalebox{0.75}{
    \begin{tikzpicture}
        \begin{axis}[ylabel=ECR($kwh/km$),xlabel=Driver Behaviour RPA,width=8cm]
    \addplot [thick,ticks=none] table [x=c1, y=c2, col sep=comma] {Data/rpa.txt};
    \addplot [name path=upper,draw=none] table [x=c1, y=c3, col sep=comma] {Data/rpa.txt};
    \addplot [name path=lower,draw=none] table [x=c1, y=c4, col sep=comma] {Data/rpa.txt};
    \addplot [fill=red!20,legend entry=\text{95\% Confidence Interval}] fill between[of=upper and lower];
    \end{axis}
    \end{tikzpicture}} \label{ecr_db} }\hspace{5pt}
    \subfigure[Energy Consumption for different trip distances]{\scalebox{0.75}{
    \begin{tikzpicture}
        \begin{axis}[ylabel=Energy Consumption ($kwh/km$),xlabel=Distance(km),width=8cm]
    \addplot [thick,ticks=none] table [x=c1, y=c2, col sep=comma] {Data/dist.txt};
    \addplot [name path=upper,draw=none] table [x=c1, y=c3, col sep=comma] {Data/dist.txt};
    \addplot [name path=lower,draw=none] table [x=c1, y=c4, col sep=comma] {Data/dist.txt};
    \addplot [fill=green!20,legend entry=\text{95\% Confidence Interval}] fill between[of=upper and lower];
    \end{axis}
    \end{tikzpicture}} \label{ecr_dist} }\\
    \subfigure[ECR for Different Positive Elevation Changes]{\scalebox{0.75}{
    \begin{tikzpicture}
    \begin{axis}[ylabel=ECR($kwh/km$),xlabel=Pos Elev Change($m$),width=8cm]
    \addplot [thick,ticks=none] table [x=c1, y=c2, col sep=comma] {Data/pos_elev.txt};
    \addplot [name path=upper,draw=none] table [x=c1, y=c3, col sep=comma] {Data/pos_elev.txt};
    \addplot [name path=lower,draw=none] table [x=c1, y=c4, col sep=comma] {Data/pos_elev.txt};
    \addplot [fill=orange!20,legend entry=\text{95\% Confidence Interval}] fill between[of=upper and lower];
    \end{axis}
    \end{tikzpicture}} \label{ecr_pe} }\hspace{5pt}
    \subfigure[ECR for Different Negative Elevation Changes]{\scalebox{0.75}{
  \begin{tikzpicture}
    \begin{axis}[yticklabel style={
        /pgf/number format/fixed,
        /pgf/number format/precision=2
},
scaled y ticks=false,ylabel=ECR($kwh/km$),xlabel=Neg Elev Change($m$),width=8cm]
    \addplot [thick,ticks=none] table [x=c1, y=c2, col sep=comma] {Data/neg_elev.txt};
    \addplot [name path=upper,draw=none] table [x=c1, y=c3, col sep=comma] {Data/neg_elev.txt};
    \addplot [name path=lower,draw=none] table [x=c1, y=c4, col sep=comma] {Data/neg_elev.txt};
    \addplot [fill=cyan!20,legend entry=\text{95\% Confidence Interval}] fill between[of=upper and lower];
    \end{axis}
    \end{tikzpicture}} \label{ecr_ne} }
    \caption{Effects of Different Factors in EV Energy Use\\\textit{ECR} : Energy Consumption Rate (\textit{watt-sec/m})}
\label{fac_eff}
\end{figure}
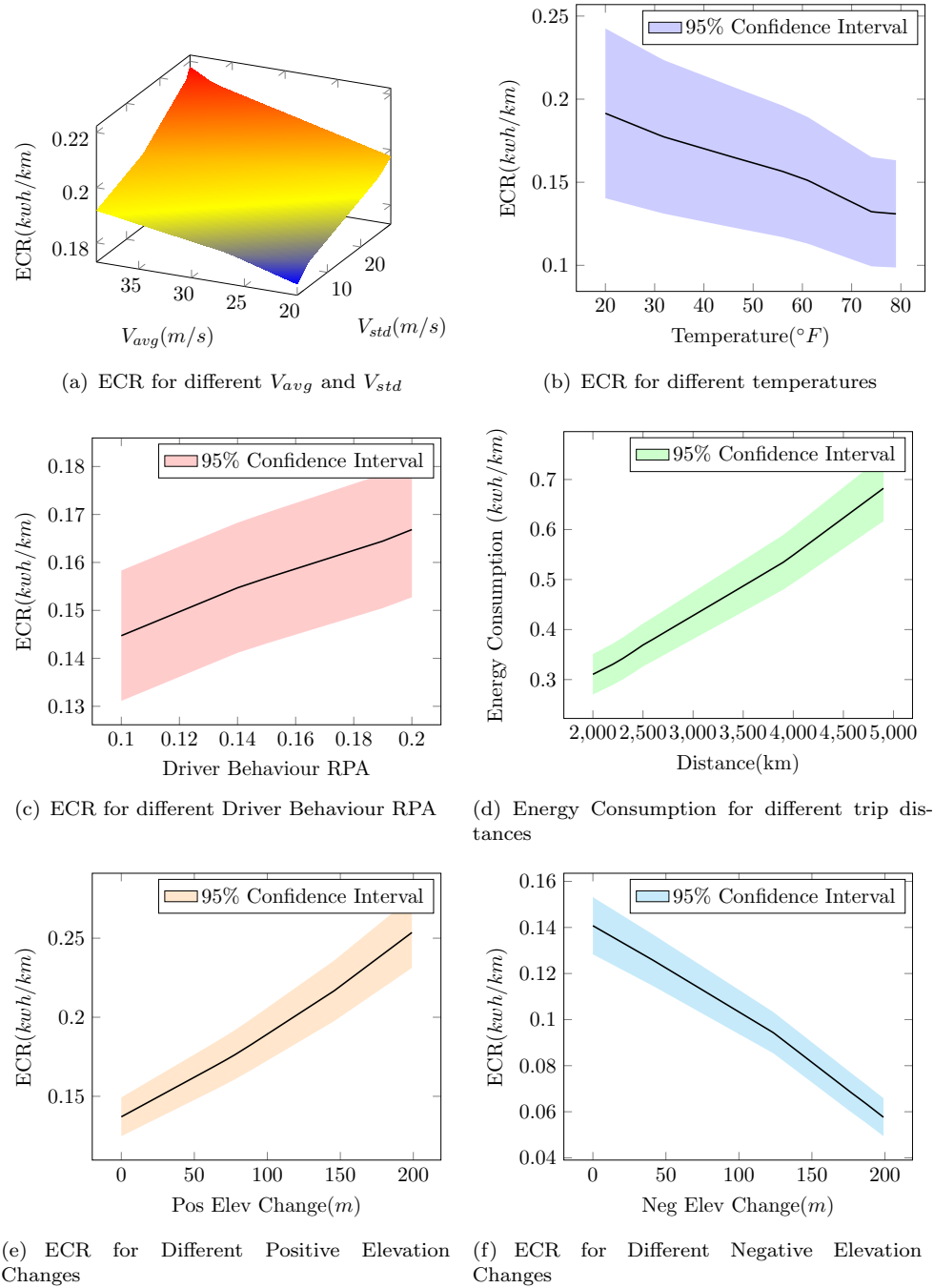

\subsection{Effects of Different Factors}

In Figure \ref{fac_eff}, we have plotted effects of diiferent factors in EV energy consumption. The plots in Figure \ref{fac_eff} show how Energy Consumption Rate (energy use per unit distance) or ECR changes with the change in various contributing factors. All the plots in Figure \ref{fac_eff} (except in Figure \ref{ecr_vdm}) show the uncertainty in energy consumption using  95\% confidence interval. Figure \ref{ecr_vdm} gives us an idea about the effects of Average Speed ($V_{avg}$) and Standard Deviation of speed ($V_{std}$) on energy consumption. As the $V_{std}$ and $V_{avg}$ increase, ECR also increases rapidly.

In Figure \ref{ecr_temp}, we can observe that increase in temperature actually reduces ECR as battery efficiency decreases in colder environment. ECR uncertainty also reduces with increase in temperature. Driver Behaviour contributes a lot in ECR. RPA (Relative Positive Acceleration) is higher for aggressive driving and it is lower for calm driving. ECR increases with the increase in RPA, which is depicted in Figure \ref{ecr_db}.

Figure \ref{ecr_dist} shows that ECR increases almost linearly with trip distance, but the uncertainty goes up as the trip gets longer. In Figure \ref{ecr_pe}, the plot shows that ECR is higher for higher positive elevation change. The effect of negative elevation change is exactly the opposite. ECR decreases slowly with increase in Negative Elevation, which is shown in Figure \ref{ecr_ne}.

\begin{figure}[htbp]
\centering
\scalebox{1.1}{
\begin{tikzpicture}
\begin{axis}
[
    xbar,
    enlarge y limits  = 0.1,
    ytick = data,
    xmin = 0,
    xmax = 0.8,
    nodes near coords,
    symbolic y coords = {Average Speed,Stddev Speed,Pos Elevation Change,Neg Elevation Change,Temperature,RPA,Average Acceleration,Average Deceleration},
  ]

\addplot coordinates {(0.08,Average Speed) (0.02,Stddev Speed)  (0.33,Pos Elevation Change) (0.24,Neg Elevation Change) (0.05,Temperature) (0.12,RPA) (0.05,Average Acceleration) (0.09,Average Deceleration)};

\end{axis}
\end{tikzpicture}}
\caption{Importance (in \%) for Different Features in EV Energy Cosnumption Estimation}
\label{imp_plot}
\end{figure}
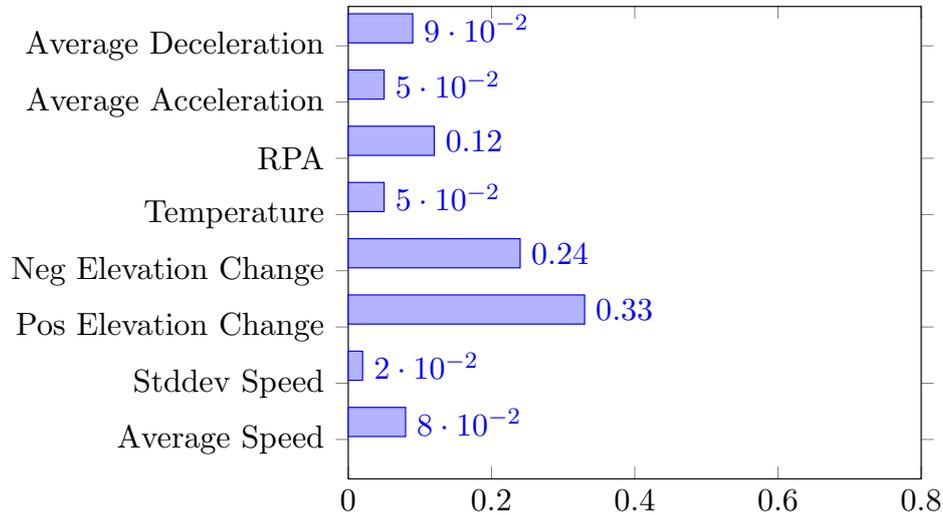

We have computed the importance of individual features to investigate how much different features impact energy consumption of EVs. Permutation feature importance \citep{breiman_no_2001} has been used to find out importance of different features. Permutation feature importance of a given feature is defined as the decrease in the model accuracy after that feature is randomly shuffled. In Figure \ref{imp_plot}, we have plotted importance of all features. It can be observed from Figure \ref{imp_plot} that Environmental Features have the highest impact on energy consumption. Positive elevation change and negative elevation change are the two most important features with importance factors 33\% and 24\% respectively. RPA is the third most dominant factor with 12\% importance. RPA is also the most important driver behaviour feature. 

\section{Limitations}
In this research study, we used only 3524 EV micro-trips to train our data-driven model. It would benefit our neural network model if trained on a wider dataset containing more trips or micro-trips. The ChargeCar dataset does not include wind speed, which can be crucial in energy estimation. The dataset also does not include rain data for all the trips. Therefore, we could not use wind speed and rain as input features for energy estimation. Battery aging or degradation impacts EV energy efficiency. However, our method has not been able to consider battery degradation as our dataset did not include that.

\section{Conclusion and Future Work}
In this research article, we have studied how energy consumption is affected by different contributing trip features. We have generated various vehicle dynamic, environmental and driver behaviour features and used those features for EV energy consumption estimation. We have proposed a probabilistic neural network with weight uncertainty for EV energy consumption estimation. The uncertainty quantification in energy consumption has the potential to reduce the range anxiety for drivers. Using energy consumption uncertainty, the EV energy consumption model can predict a confidence interval for the energy consumption of a trip, which the drivers can use to plan the trip route, charging stops and recharging amounts. 

The main results can be summarized as follows:
\begin{itemize}
    \item The MAPE error of our probabilistic proposed is 9.3\%. Our proposed probabilistic model with weight uncertainty reduces the energy estimation error by 19.8\% compared to the probabilistic model without weight uncertainty. The weight uncertainty has helped our proposed model avoid overfitting and increase the model's robustness to noise.
    \item This study has observed that probabilistic models outperform deterministic models and our proposed probabilistic model reduces the energy estimation error by 34.5\% compared to deterministic MLP NN model.
    \item We have computed and analysed the importance of different features in EV energy consumption. Our study shows that using relative positive acceleration (RPA), average acceleration and deceleration as driver behaviour features significantly improves the model's accuracy. The use of driver behaviour features reduces the estimation error by 27.9\%. Relative positive acceleration (RPA) is the most important driver behaviour feature with an importance factor of 12\%.
    \item Our results show that using positive and negative elevation change as two separate features is crucial, as the negative elevation change is beneficial in finding and computing energy regeneration. Positive and negative elevation change are the two most important input features in EV trip energy consumption estimation with importance factors 33\% and 24\% repectively. 
\end{itemize}

In our future work, we will consider other important environmental and physical factors as input features for energy estimation that we could not include in this model. It would be interesting to observe the model accuracy after including wind speed and rain. Battery degradation can be a significant contributing factor in energy consumption, which we would like to include in future. We are also willing to validate our model on wider EV energy consumption datasets. As EV trip data is a time series data, future work should explore the applicability of sequential deep learning models (e.g. recurrent neural networks (RNN), long short-term memory RNN etc.) in EV trip energy consumption estimation. Possible future work also includes building an EV routing planner and EV charging-discharging scheduling system using energy consumption uncertainty.

\section{Conflicts of Interest}

The authors declare that they have no conflicts of interest.

\section{CRediT Authorship Contribution Statement}

\textbf{Ayan Maity} : Conceptualization, Methodology, Data Curation, Software, Validation, Visualization, Writing - Original Draft, Writing - Review and Editing. \textbf{Sudeshna Sarkar} :  Conceptualization, Methodology, Supervision, Writing - Review and Editing.

\bibliographystyle{tfcad}
\bibliography{Bibliography}

\end{document}